\documentclass[11pt,a4paper]{article}
\usepackage[hyperref]{emnlp2020}
\usepackage{times}
\usepackage{latexsym}

\usepackage{booktabs}
\usepackage{color}
\usepackage{tikz-dependency}
\usepackage{amsfonts}
\usepackage{amsmath}
\usepackage{multirow}
\usepackage{makecell}
\usepackage{pifont}

\newcommand{\xmark}{\ding{55}}
\usepackage[noend]{algpseudocode}
\usepackage{algorithmicx,algorithm}

\usepackage{microtype}

\aclfinalcopy 
\def\aclpaperid{704} 

\title{A Multi-task Learning Framework for Opinion Triplet Extraction}

\author{Chen Zhang\textsuperscript{1}, Qiuchi Li\textsuperscript{2}, Dawei Song\textsuperscript{1}\Thanks{ Dawei Song is the corresponding author.}, Benyou Wang\textsuperscript{2} \\
  \textsuperscript{1} Beijing Institute of Technology, Beijing, China. \\
  \textsuperscript{2} University of Padova, Padova, Italy. \\
   \texttt{\{czhang,dwsong\}@bit.edu.cn}, \texttt{\{qiuchili,wang\}@dei.unipd.it} \\}

\date{}

\begin{document}
\maketitle
\begin{abstract}
The state-of-the-art Aspect-based Sentiment Analysis (ABSA) approaches are mainly based on either detecting aspect terms and their corresponding sentiment polarities, or co-extracting aspect and opinion terms. However, the extraction of aspect-sentiment pairs lacks opinion terms as a reference, while co-extraction of aspect and opinion terms would not lead to meaningful pairs without determining their sentiment dependencies. To address the issue, we present a novel view of ABSA as an opinion triplet extraction task, and propose a multi-task learning framework to jointly extract aspect terms and opinion terms, and simultaneously parses sentiment dependencies between them with a biaffine scorer. At inference phase, the extraction of triplets is facilitated by a triplet decoding method based on the above outputs. We evaluate the proposed framework on four SemEval benchmarks for ASBA. The results demonstrate that our approach significantly outperforms a range of strong baselines and state-of-the-art approaches.\footnote{Code and datasets for reproduction are available at \href{https://github.com/GeneZC/OTE-MTL}{\texttt{https://github.com/GeneZC/OTE-MTL}}.}  
\end{abstract}

\section{Introduction}

Aspect-based sentiment analysis (ABSA), also termed as Target-based Sentiment Analysis in some literature~\cite{liu2012sentiment}, is a fine-grained sentiment analysis task. It is usually formulated as detecting aspect terms and sentiments expressed in a sentence towards the aspects~\cite{li2019unified,he2019interactive,luo2019doer,hu2019open}. This type of formulation is referred to as \emph{aspect-sentiment pair extraction}.   Meanwhile, there exists another type of approach to ABSA,  referred to as \emph{aspect-opinion co-extraction}, which focuses on jointly deriving aspect terms (a.k.a. opinion targets) and opinion terms (a.k.a. opinion expressions) from sentences, yet without figuring out their sentiment dependencies~\cite{wang2017coupled,li2018aspect}. The compelling performances of both directions illustrate a strong dependency between aspect terms, opinion terms and the expressed sentiments.

\begin{figure}
    \centering
    \resizebox{.47\textwidth}{!}{
    \begin{tabular}{rc}
    \toprule
    Example sentence: & \makecell{The \textcolor{blue}{atmosphere} is \textcolor{red}{attractive} ,\\ but a little \textcolor{red}{uncomfortable} .} \\
    \midrule
    Aspect-sentiment pair extraction : & \makecell{[(\textcolor{blue}{atmosphere}, positive),\\  (\textcolor{blue}{atmosphere}, negative)]} \\
    \midrule
    Aspect-opinion co-extraction : &  \makecell{[\textcolor{blue}{atmosphere}, \textcolor{red}{attractive},\\ \textcolor{red}{uncomfortable}]} \\
    \midrule
    Opinion triplet extraction : &  \makecell{[(\textcolor{blue}{atmosphere}, \textcolor{red}{attractive}, positive),\\ (\textcolor{blue}{atmosphere}, \textcolor{red}{uncomfortable}, negative)]} \\
    \bottomrule
    \end{tabular}
    }
    \caption{Differences among aspect-sentiment pair extraction, aspect-opinion co-extraction, and opinion triplet extraction. Words in \textcolor{blue}{blue} are aspect terms. Words in \textcolor{red}{red} are opinion terms. [ ] denotes a set of extracted patterns, and ( ) denotes an extracted pattern.}
    \label{fig1}
\end{figure}

This motivates us to put forward a new perspective for ABSA as joint extraction of aspect terms, opinion terms and sentiment polarities,\footnote{For simplicity, these four concepts are hereafter referred to as aspect, opinion, sentiment, and triplet, respectively.} in short \textit{opinion triplet extraction}. An illustrative example of differences among aspect-sentiment pair extraction, aspect-opinion co-extraction, and opinion triplet extraction is given in Figure~\ref{fig1}. Opinion triplet extraction can be viewed as an integration of aspect-sentiment pair extraction and aspect-opinion co-extraction, by taking into consideration their complementary nature. It brings in two-fold advantages: (1) the opinions can boost the expressive power of models and help better determine aspect-oriented sentiments; (2) the sentiment dependencies between aspects and opinions can bridge the gap of how sentiment decisions are made and further promote interpretability of models. 

There is some prior research with a similar viewpoint. \newcite{peng2019knowing} proposes to extract opinion tuples, i.e., (aspect-sentiment pair, opinion)s,\footnote{To some extent, opinion triplet extraction aims at solving the same task (named aspect sentiment triplet extraction) as they does regardless of the minor difference.} by first jointly extracting aspect-sentiment pairs and opinions by two sequence taggers, in which sentiments are attached to aspects via unified tags,\footnote{An aspect tag set \{\texttt{B}, \texttt{I}, \texttt{O}\} and a sentiment tag set \{\texttt{NEU}, \texttt{NEG}, \texttt{POS}\} are unified into the aspect-sentiment tag set \{\texttt{B-NEU}, \texttt{I-NEU}, \texttt{B-NEG}, \texttt{I-NEG}, \texttt{B-POS}, \texttt{I-POS}, \texttt{O}\}. Here, \texttt{B}, \texttt{I}, and \texttt{O} indicate begin, inside, and outside of a span. And \texttt{NEU}, \texttt{NEG}, and \texttt{POS} are neutral, negative, and positive.} and then pairing the extracted aspect-sentiments and
opinions by an additional classifier. Despite of remarkable performance the approach has achieved, two issues need to be addressed.

The first issue arises from the prediction of aspects and sentiments with a set of unified tags thus degrading the sentiment dependency parsing process to a binary classification. As is discussed in prior studies on aspect-sentiment pair extraction~\cite{he2019interactive,luo2019doer,hu2019open}, although the concerned framework with unified tagging scheme is theoretically  elegant and mitigates the computational cost, it is insufficient to model the interaction between the aspects and sentiments~\cite{he2019interactive,luo2019doer}.

\begin{figure}
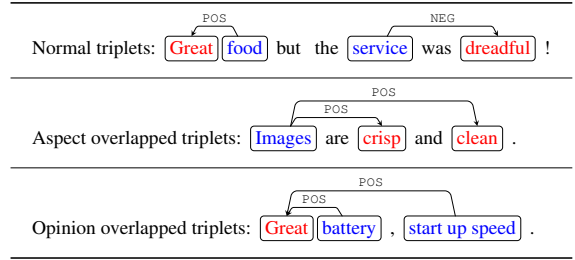

    \centering
    \resizebox{.47\textwidth}{!}{
    \begin{tabular}{l}
    \toprule
    \begin{dependency}[text only label, label style={above}]
    \begin{deptext}[column sep=.1cm]
    Normal triplets: \& \textcolor{red}{Great} \& \textcolor{blue}{food} \& but \& the \& \textcolor{blue}{service} \& was \& \textcolor{red}{dreadful} \& ! \\ 
    \end{deptext} 
    \depedge[edge height=.2cm]{3}{2}{\texttt{POS}} \depedge[edge height=.2cm]{6}{8}{\texttt{NEG}}
    \wordgroup{1}{2}{2}{op1}
    \wordgroup{1}{3}{3}{tg1}
    \wordgroup{1}{8}{8}{op1}
    \wordgroup{1}{6}{6}{tg1}
    \end{dependency} \\
    \midrule
    \begin{dependency}[text only label, label style={above}, edge height=.3cm]
    \begin{deptext}[column sep=.1cm]
    Aspect overlapped triplets: \& \textcolor{blue}{Images} \& are \& \textcolor{red}{crisp} \& and \& \textcolor{red}{clean} \& . \\
    \end{deptext} 
    \depedge[edge height=.17cm]{2}{4}{\texttt{POS}} \depedge[edge height=.5cm]{2}{6}{\texttt{POS}}
    \wordgroup{1}{4}{4}{op1}
    \wordgroup{1}{2}{2}{tg1}
    \wordgroup{1}{6}{6}{op2}
    \end{dependency} \\
    \midrule
    \begin{dependency}[text only label, label style={above}, edge height=.3cm]
    \begin{deptext}[column sep=.1cm]
    Opinion overlapped triplets: \& \textcolor{red}{Great} \& \textcolor{blue}{battery} \& , \& \textcolor{blue}{start up speed} \& . \\
    \end{deptext} 
    \depedge[edge height=.17cm]{3}{2}{\texttt{POS}} \depedge[edge height=.5cm]{5}{2}{\texttt{POS}}
    \wordgroup{1}{3}{3}{tg1}
    \wordgroup{1}{2}{2}{op1}
    \wordgroup{1}{5}{5}{tg2}
    \end{dependency} \\
    \bottomrule
    \end{tabular}
    }
    \caption{Categories of triplets. Spans in \textcolor{blue}{blue} are aspects and spans in \textcolor{red}{red} are opinions. Arcs indicate sentiment dependencies and are always directed from an aspect to opinion.}
    \label{fig2}
\end{figure}

Secondly, the coupled aspect-sentiment formalization disregards the importance of their interaction with opinions. Such interaction has been shown important to handle the overlapping circumstances where different triplet patterns share certain elements, in other triplet extraction-based tasks such as relation extraction~\cite{fu2019graphrel}. To show why triplet interaction modelling is crucial, we divide triplets into three categories, i.e., aspect overlapped, opinion overlapped, and normal ones. Examples of these three kinds of triplets are shown in Figure~\ref{fig2}. We can observe that two triplets tend to have the same sentiment if they share the same aspect or opinion. Hence, modelling triplet interaction shall benefit the ASBA task, yet it can not be explored with the unified aspect-sentiment tags in which sentiments have been attached to aspects without considering the overlapping cases.

To circumvent the above issues, we propose a multi-task learning framework for opinion triplet extraction, namely OTE-MTL, to jointly detect aspects, opinions, and sentiment dependencies. On one hand, the aspects and opinions can be extracted with two independent heads in the multi-head architecture we propose. On the other hand, we decouple sentiment prediction from aspect extraction. Instead, we employ a sentiment dependency parser as the third head, to predict word-level sentiment dependencies, which will be utilized to further decode span-level\footnote{The aspects and opinions are usually spans over several words in the sentence} dependencies when incorporated with the detected aspects and opinions. In doing so, we expect to alleviate issues brought by the unified tagging scheme. Specifically, we exploit sequence tagging strategies~\cite{lample2016neural} for extraction of aspects and opinions, whilst taking advantage of a biaffine scorer~\cite{dozat2017deep} to obtain word-level sentiment dependencies. Additionally, since these task-heads are jointly trained, the learning objectives of aspect and opinion extraction could be considered as regularization applied on the sentiment dependency parser. In this way, the parser is learned with aspect- and opinion-aware constraints, therefore fulfilling the demand of triplet interaction modelling. Intuitively, if we are provided with a sentence containing two aspects but only one opinion (e.g., the third example in Figure~\ref{fig2}), we can identify triplets with overlapped opinion thereby.

Extensive experiments are carried out on four SemEval benckmarking data collections for ABSA. Our framework are compared with a range of state-of-the-art approaches. The results demonstrate the effectiveness of our overall framework and individual components within it. A further case study shows that how our model better handles overlapping cases.
\section{Proposed Framework}

\begin{figure*}
    \centering
    \includegraphics[width=.7\textwidth]{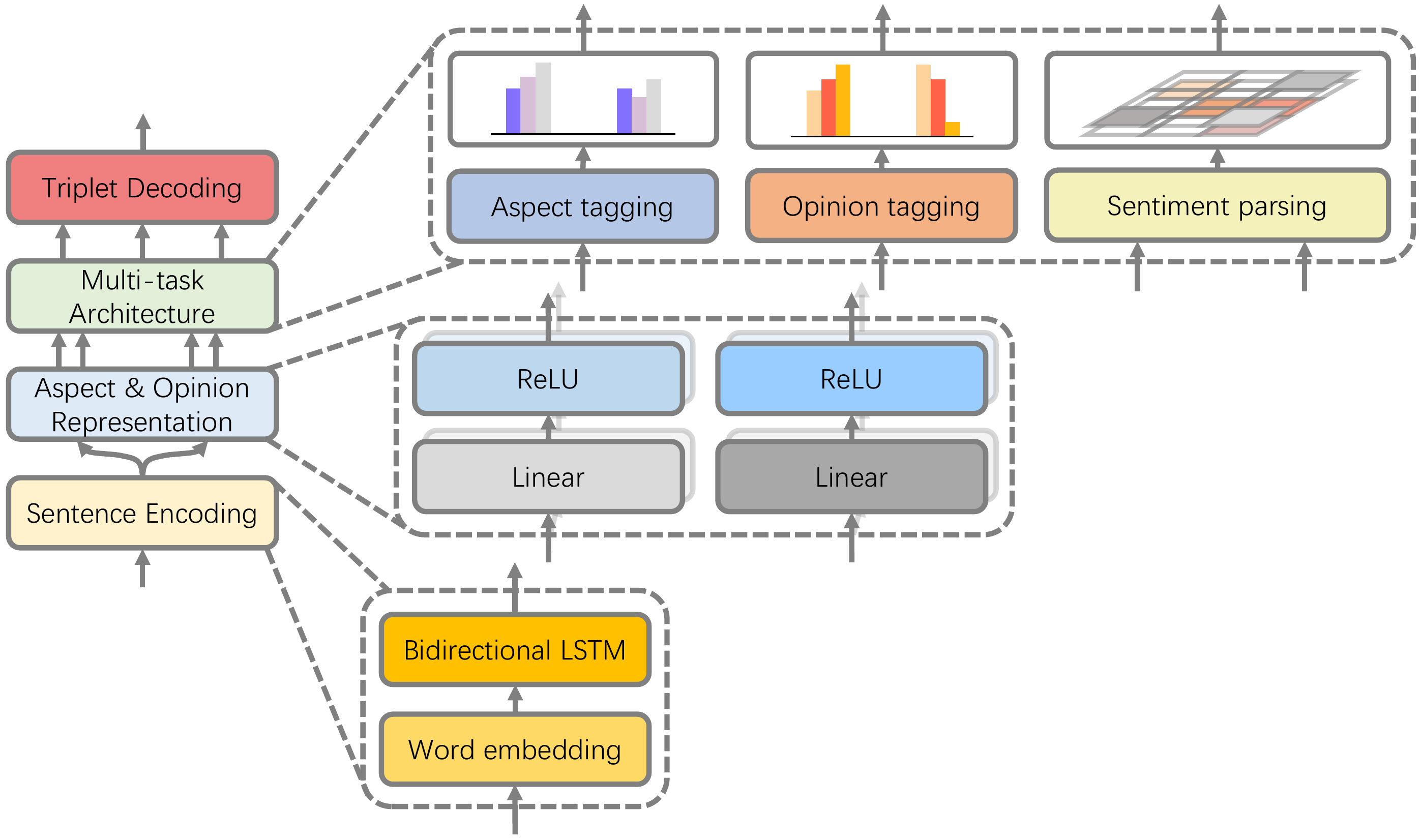}
    \caption{An overview of our proposed framework.}
    \label{fig3}
\end{figure*}

\subsection{Problem Formulation}

Given an input sentence $S=\{w_i\}_{i=1}^{|S|}$, our model aims to output a set of triplets $T=\{t_j\}_{j=1}^{|T|}$, where $|S|$, $|T|$ are the lengths of the sentence and the triplet set, respectively. A triplet $t_j$ consists of three elements, i.e., $[m_{j}^{(ap)},m_{j}^{(op)},m_{j}^{(st)}]$, which separately stand for aspect span, opinion span, and sentiment. While the aspects and opinions are usually spans over several words in the sentence, we simplify the notation with the start position (denoted as $sp$) and end position (denoted as $ep$) of a span. Accordingly, $m_{j}^{(ap)}$ and $m_{j}^{(op)}$ can be represented as $(sp_{j}^{(ap)},ep_{j}^{(ap)})$ and $(sp_{j}^{(op)},ep_{j}^{(op)})$. Thus, the problem is formulated as finding a function $\mathcal{F}$ that accurately maps the sentence $S=\{w_i\}_{i=1}^{|S|}$ onto a triplet set $T=\{t_j\mid t_j=[(sp_{j}^{(ap)},ep_{j}^{(ap)}),(sp_{j}^{(op)},ep_{j}^{(op)}),m_{j}^{(st)}]\}_{j=1}^{|T|}$.

\subsection{The OTE-MTL Framework}

Our proposed OTE-MTL framework folds the triplet extraction process into two stages, i.e., prediction stage and decoding stage. An overview of our framework is presented in Figure~\ref{fig3}. The prediction stage is parameterized by neural models and thus is trainable. It builds upon a sentence encoding module based on word embedding and a bidirectional LSTM structure, to learn an abstract representation of aspects and opinions. Underpinned by the abstract representation, there are three core components, accounting for three subgoals, i.e., aspect tagging, opinion tagging, and word-level sentiment dependency parsing. After the aspects, opinions and word-level dependencies have been detected, a decoding stage is then carried out to produce triplets based on heuristic rules.

\subsection{Sentence Encoding}

Context awareness is crucial for sentence encoding, i.e., encoding a sentence into a sequence of vectors. Hence, we adopt a bidirectional Long Short-term Memory network (LSTM)~\cite{hochreiter1997long} as our sentence encoder, owing to the context modelling capability of LSTMs. In order to encode the input sentence, we first embed each word in a sentence to a low-dimensional vector space~\cite{bengio2003neural} with pre-trained word embeddings\footnote{In our experiments, GloVe vectors~\cite{pennington2014glove} are used.}. With the embedded word representations $E=\{\mathbf{e}_i\mid \mathbf{e}_i\in \mathbb{R}^{d_e}\}_{i=1}^{|S|}$, the bidirectional LSTM is employed to attain contextualized representations of words  $H=\{\mathbf{h}_i\mid \mathbf{h}_i\in \mathbb{R}^{2d_h}\}_{i=1}^{|S|}$ by the following operation:
\begin{equation}
    \mathbf{h}_i=[\overrightarrow{\mathrm{LSTM}}(\mathbf{e}_i)\oplus\overleftarrow{\mathrm{LSTM}}(\mathbf{e}_i)]
\end{equation}
where $d_e$ and $d_h$ denote the dimensionality of a word embedding and a hidden state from an uni-directional LSTM, while $\overrightarrow{\mathrm{LSTM}}(\cdot)$ and $\overleftarrow{\mathrm{LSTM}}(\cdot)$ stand for forward and backward LSTM, respectively. $\oplus$ means vector concatenation. 

\subsection{Aspect and Opinion Representation}

We then extract the aspect- and opinion-specific features from the encoded hidden states, by applying dimension-reducing linear layers and nonlinear functions, rather than directly feeding the hidden states into the next components, for two reasons. First, the hidden states might contain superfluous information for follow-on computations, potentially causing a risk of overfitting. Second, such operations are expected to strip away irrelevant features for aspect tagging and opinion tagging. The computation process is formulated as below:
\begin{gather}
    \mathbf{r}_i^{(ap)}=g(\mathbf{W}_r^{(ap)}\mathbf{h}_i+\mathbf{b}_r^{(ap)}) 
    \label{eq2}\\ 
    \mathbf{r}_i^{(op)}=g(\mathbf{W}_r^{(op)}\mathbf{h}_i+\mathbf{b}_r^{(op)})
    \label{eq3}
\end{gather}
where $\mathbf{r}_i^{(ap)}\in \mathbb{R}^{d_r}$ and $\mathbf{r}_i^{(op)}\in \mathbb{R}^{d_r}$ are aspect and opinion representations, $d_r$ is the dimensionality of the representation. $\mathbf{W}_r^{(ap)}$, $\mathbf{W}_r^{(op)}\in \mathbb{R}^{d_r\times 2d_h}$ and $\mathbf{b}_r^{(ap)}$, $\mathbf{b}_r^{(op)}\in \mathbb{R}^{d_r}$ are learnable weights and biases. Here, $g(\cdot)$ is a nonlinear function, which is ReLU, i.e., $\max(\cdot,0)$, in our case.

Note that above representations are prepared for tagging. Likewise, we obtain another set of representations $\mathbf{r}_i^{(ap)\prime}, \mathbf{r}_i^{(op)\prime}\in \mathbb{R}^{d_r}$ for sentiment parsing, following the same procedure as Equation~\ref{eq2} and~\ref{eq3} but with different parameters.

\subsection{Multi-task Architecture}

The multi-task architecture includes two parts: aspect and opinion tagging, and word-level sentiment dependency parsing. 

\noindent \textbf{Aspect and Opinion Tagging.} Following the \{\texttt{B}, \texttt{I}, \texttt{O}\} tagging scheme, we tag each word in the sentence with two taggers, i.e., one tagger for aspect, and the other for opinion. In particular, we receive two series of distributions over \{\texttt{B}, \texttt{I}, \texttt{O}\} tags $\mathbf{p}_i^{(ap)}$ and $\mathbf{p}_i^{(op)}\in \mathbb{R}^{3}$ through:
\begin{gather}
    \mathbf{p}_i^{(ap)}=\mathrm{softmax}(\mathbf{W}_t^{(ap)}\mathbf{r}_i^{(ap)}+\mathbf{b}_t^{(ap)}) \\ 
    \mathbf{p}_i^{(op)}=\mathrm{softmax}(\mathbf{W}_t^{(op)}\mathbf{r}_i^{(op)}+\mathbf{b}_t^{(op)})
\end{gather}
where $\mathbf{W}_t^{(ap)}$, $\mathbf{W}_t^{(op)}\in \mathbb{R}^{3\times d_r}$ and $\mathbf{b}_t^{(ap)}$, $\mathbf{b}_t^{(op)}\in \mathbb{R}^{3}$ are trainable parameters.

Accordingly, we can deduce the loss function, typically cross entropy with categorical distribution, for tagging as:
\begin{equation}
\begin{aligned}
    \mathcal{L}_{tag}=-\frac{1}{|S|}\sum_i\sum_k \hat{\mathbf{p}}_{i,k}^{(ap)}\mathrm{log}(\mathbf{p}_{i,k}^{(ap)})\\ 
    -\frac{1}{|S|}\sum_i\sum_k \hat{\mathbf{p}}_{i,k}^{(op)}\mathrm{log}(\mathbf{p}_{i,k}^{(op)}) 
\end{aligned}
\end{equation}
where $\hat{\mathbf{p}}_{i}^{(ap)}$ and $\hat{\mathbf{p}}_{i}^{(op)}$ respectively denote the ground truth aspect and opinion tag distributions of each word, and $k$ is an enumerator over each item in a categorical distribution.

\noindent \textbf{Word-level Sentiment Dependency Parsing.} There are $|S|^2$ possible word pairs (including self-pairing cases) in each sentence and we intend to determine dependency type of every word pair. The set of dependency types is defined as \{\texttt{NEU}, \texttt{NEG}, \texttt{POS}, \texttt{NO-DEP}\}, so as to address all kinds of dependencies. Here, \texttt{NO-DEP} denotes no sentiment dependency. In addition, inspired by the table filling methods~\cite{miwa2014modeling,bekoulis2018joint}, sentiment dependencies are considered only for a pair of words that are exactly the last word of an aspect and the last word of an opinion in a triplet. Recall the example sentence ``\textit{Great battery, start up speed.}''. For the triplet (\textit{start up speed}, \textit{great}, \texttt{POS}), the sentiment dependency is simplified to (\textit{speed}, \textit{great}, \texttt{POS}). As such, the learning redundancy for the parser is much reduced, while the span-level sentiment dependency is still available when it is combined with extracted aspect and opinion spans.

We utilize a biaffine scorer to capture the interaction of two words in each word pair, due to its proven expressive power in syntactic dependency parsing~\cite{dozat2017deep}. The score assignment to each word pair is as below:
\begin{equation}
\begin{aligned}
    \tilde{\mathbf{s}}_{i,j,k}=[\mathbf{W}^{(k)}\mathbf{r}_i^{(ap)\prime}+\mathbf{b}^{(k)}]^{\top}\mathbf{r}_j^{(op)\prime}\\
    =[\mathbf{W}^{(k)}\mathbf{r}_i^{(ap)\prime}]^{\top}\mathbf{r}_j^{(op)\prime}+{\mathbf{b}^{(k)}}^{\top}\mathbf{r}_j^{(op)\prime}
    \label{eq7}
\end{aligned}
\end{equation}
where $\tilde{\mathbf{s}}_{i,j,k}$ stands for score of the $k$-th dependency type for a word pair $(w_i,w_j)$. $\mathbf{W}^{(k)}$ and $\mathbf{b}^{(k)}$ are trainable weight and bias for producing the $k$-th score, respectively. Moreover, we use $\mathbf{s}_{i,j}$ to indicate a softmax-normalized vector of scores, which contains probabilities of all dependency types for the word pair $(w_i,w_j)$:
\begin{equation}
    \mathbf{s}_{i,j,k}=\mathrm{softmax}(\tilde{\mathbf{s}}_{i,j,k})
\end{equation}

As observed from the factorization in Equation~\ref{eq7}, conceptually the biaffine scorer can not only model the likelihood of $w_i$ receiving $w_j$ as a dependent of a specific type (the first term), but also include the prior probability of $w_j$ being a dependent of such type (the second term). When it is implemented, the scorer is essentially an affine transform followed by matrix multiplication.

Thereafter, the loss function for word-level sentiment dependency parsing is a cross entropy function given below:
\begin{equation}
    \mathcal{L}_{dep}=-\frac{1}{|S|^2}\sum_{(i,j)}\sum_k\hat{\mathbf{s}}_{i,j,k}\mathrm{log}(\mathbf{s}_{i,j,k})
\end{equation}
where $\hat{\mathbf{s}}_{i,j}$ is the ground-truth dependency distribution for each word pair $(w_i,w_j)$.

\noindent \textbf{Overall Learning Objective.} Ultimately, we can conduct joint training of the multi-task learning framework with the following objective:
\begin{equation}
    \min_{\mathbf{\theta}}\mathcal{L}=\min_{\mathbf{\theta}}\mathcal{L}_{tag}+\alpha\mathcal{L}_{dep}+\gamma ||\theta||_2
\end{equation}
where $\alpha$ is a trade-off term to balance the learning between tagging and sentiment dependency parsing. $\mathbf{\theta}$ stands for trainable parameters. $||\mathbf{\theta}||_2$ and $\gamma$ are $L_2$ regularization of $\mathbf{\theta}$ and a controlling term, respectively.

\subsection{Triplet Decoding}

Upon obtaining the extracted aspects, opinions, and word-level sentiment dependencies, we conduct a triplet decoding process using heuristic rules. Basically, we view the sentiment dependencies resulted from the biaffine scorer as pivots, and carry out a reverse-order traverse on tags generated by the aspect and opinion taggers. 

For example, from word sequence ``\textit{Great battery , start up speed .}'', we get aspect tags \{\texttt{O}, \texttt{B}, \texttt{O}, \texttt{B}, \texttt{I}, \texttt{I}, \texttt{O}\}, opinion tags \{\texttt{B}, \texttt{O}, \texttt{O}, \texttt{O}, \texttt{O}, \texttt{O}, \texttt{O}\}, and a word-level sentiment dependency, which is represented in index form, (6, 1, \texttt{POS}). The yielded sentiment dependency typically means that the last word of aspect is the 6-th word (\textit{speed}), the last word of opinion is the 1-th word (\textit{Great}), and they together form a positive sentiment. The traverse is conducted based on the aspect and opinion index (pivots) and the word sequence following stop-on-non-\texttt{I} criterion. And the final output should be [(4, 6), (1, 1), \texttt{POS}]. Details of the algorithm is shown in~\ref{alg1}.

\begin{algorithm}
    \caption{Decoding w/ stop-on-non-\texttt{I} criterion.}
    \hspace*{0.02in} {\bf Input:}
    aspect tags $\{g^{(ap)}_i\}_{i=1}^n$, opinion tags $\{g^{(op)}_i\}_{i=1}^n$, sentiment dependency $(j,k,p)$.\\
    \hspace*{0.02in} {\bf Output:}
    triplet $t$
    \begin{algorithmic}[1]
    \State $j^{\prime}\leftarrow j$
    \While{$g^{(ap)}_{j^{\prime}}$ is \texttt{I}} $\lhd$ \textcolor{gray}{stop on \texttt{B} and \texttt{O}.}
    \State $j^{\prime}\leftarrow j^{\prime}-1$
    \If{$j^{\prime}\leq 0$} $\lhd$ \textcolor{gray}{or exceeding boundary.}
    \State break
    \EndIf
    \EndWhile
    \State $k^{\prime}\leftarrow k$
    \While{$g^{(op)}_{k^{\prime}}$ is \texttt{I}}
    \State $k^{\prime}\leftarrow k^{\prime}-1$
    \If{$k^{\prime}\leq 0$}
    \State break
    \EndIf
    \EndWhile
    \State $t\leftarrow [(j^{\prime},j),(k^{\prime},k),p]$
    \end{algorithmic}
    \label{alg1} 
\end{algorithm}

\section{Experimental Setup}

\subsection{Datasets and Evaluation Metrics}

We conduct experiments on three datasets  in the ``restaurant'' domain from SemEval 2014, 2015 and 2016~\cite{pontiki2014semeval,pontiki2015semeval,pontiki2016semeval}, and one dataset in the ``laptop'' domain from SemEval 2014. Hereafter, we will refer to them as \textsc{Rest14}, \textsc{Rest15}, \textsc{Rest16}, and \textsc{Laptop14} respectively. Since they are originally annotated with aspects and sentiments only, we additionally adopt annotations of opinion terms from~\citet{wang2017coupled} and~\citet{peng2019knowing}. Each dataset is split to three subsets, namely, training set, validation set, and test set. The statistics of these datasets are shown in Table~\ref{tab1}. It is worth noting that, in~\cite{peng2019knowing}, the opinion overlapped triplets (in short OOTs) are removed from all four datasets in the preprocessing step. However, these cases are preserved in our setting. A key observation from the statistics is that there are large amounts of overlapping cases in the datasets, on average accounting for 24.2\% of the total number of triplets across all four datasets. This phenomenon suggests the need and significance of triplet interaction modelling.

Moreover, we adopt precision, recall, and micro F1-measure as our evaluation metrics for triplet extraction. Only exactly matched triplets, i.e., with all of the aspect, opinion and sentiment matched against gold standards, are viewed as true positives during evaluation. All results are reported by averaging 10 runs with random initialization. Paired t-test is used to examine statistical significance of the results.

\begin{table}
    \centering
    \resizebox{.47\textwidth}{!}{
    \begin{tabular}{cccccc} 
    \toprule
    \multicolumn{2}{c}{Dataset}       & \# sentence & \# triplet & \makecell[c]{\# sentence\\ w/ overlap} & \makecell[c]{\# triplet\\ w/ overlap}  \\
    \midrule
    \multirow{3}{*}[-.2cm]{\textsc{Rest14}}   & train & 1300       & 2409      & 437                    & 578                    \\
    \cmidrule{2-6}
                              & val.  & 323        & 590       & 92                     & 147                    \\
    \cmidrule{2-6}
                              & test  & 496        & 1014      & 193                    & 389                    \\
    \midrule
    \multirow{3}{*}[-.2cm]{\textsc{Rest15}}   & train & 593        & 977       & 151                    & 189                    \\
    \cmidrule{2-6}
                              & val.  & 148        & 160       & 42                     & 62                     \\
    \cmidrule{2-6}
                              & test  & 318        & 479       & 68                     & 71                     \\
    \midrule
    \multirow{3}{*}[-.2cm]{\textsc{Rest16}}   & train & 842        & 1370      & 208                    & 256                    \\
    \cmidrule{2-6}
                              & val.  & 210        & 334       & 52                     & 61                     \\
    \cmidrule{2-6}
                              & test  & 320        & 507       & 77                     & 120                    \\
    \midrule
    \multirow{3}{*}[-.2cm]{\textsc{Laptop14}} & train & 920        & 1451      & 263                    & 365                    \\
    \cmidrule{2-6}
                              & val.  & 228        & 380       & 80                     & 101                    \\
    \cmidrule{2-6}
                              & test  & 339        & 552       & 103                    & 140                    \\
    \bottomrule
    \end{tabular}
    }
    \caption{Statistics of datasets. Sentence w/ overlap means sentence containing overlapped triplets and triplet w/ overlap denotes triplet that overlaps with other triplets.}
    \label{tab1}
\end{table}

\subsection{Implementation Details}

In our experiments, the word embeddings are initialized with pretrained GloVe word vectors~\cite{pennington2014glove}. The dimensionalities of embeddings $d_e$, hidden states $d_h$, aspect and opinion representations $d_r$ are set to 300, 300, 100, respectively. The trade-off term in learning objective, i.e., $\alpha$, is set to be 1. The coefficient for $L_2$ regularization, i.e., $\gamma$, is 10\textsuperscript{-5}. Dropout is applied on embeddings to avoid overfitting and the drop rate is 0.5. The learning rate during training is 10\textsuperscript{-3} while the batch size is 32. All the parameters are initialized with uniform distribution and optimized with the Adam optimizer. Besides, we set a patience number 5, so that we could stop the learning process early if there is no further performance improvement on validation set. 

\begin{table*}[ht]
    \centering
    \resizebox{\textwidth}{!}{
    \begin{tabular}{ccccccccccccc}
    \toprule
    \multirow{2}{*}[-.1cm]{Model} & \multicolumn{3}{c}{\textsc{Rest14}} & \multicolumn{3}{c}{\textsc{Rest15}} & \multicolumn{3}{c}{\textsc{Rest16}} & \multicolumn{3}{c}{\textsc{Laptop14}} \\
    \cmidrule{2-13}
    & pre. & rec. & f1. & pre. & rec. & f1. & pre. & rec. & f1. & pre. & rec. & f1. \\
    \midrule
    RENANTE+\textsuperscript{\dag *} & 30.90 & 38.30 & 34.20 & 29.40 & 26.90 & 28.00 & 27.10 & 20.50 & 23.30 & 23.10 & 17.60 & 20.00\\
    CMLA+\textsuperscript{\dag *} & 38.80 & 47.10 & 42.50 & 34.40 & 37.60 & 35.90 & 43.60 & 39.80 & 41.60 & 31.40 & 34.60 & 32.90\\
    Unified+\textsuperscript{\dag *} & 43.83 & 62.38 & 51.43 & 43.34 & 50.73 & 46.69 & 38.19 & 53.47 & 44.51 & 42.25 & 42.78 & 42.47\\
    Pipeline\textsuperscript{\dag *} & 42.29 & 64.07 & 50.90 & 40.97 & 54.68 & 46.79 & 46.76 & 62.97 & 53.62 & 40.40 & 47.24 & 43.50\\
    OTE-MTL (ours)\textsuperscript{*} & 66.04 & 56.25 & \textbf{60.62}\textsuperscript{\ddag} & 57.51 & 43.96 & \textbf{49.76}\textsuperscript{\ddag} & 64.68 & 54.97 & \textbf{59.36}\textsuperscript{\ddag} & 50.52 & 39.71 & \textbf{44.31}\textsuperscript{\ddag}\\
    \midrule
    CMLA-MTL & 43.24 & 44.95 & 43.97 & 35.87 & 39.85 & 37.55 & 44.22 & 46.43 & 45.01 & 33.61 & 36.11 & 34.68\\
    HAST-MTL & 58.97 & 46.75 & 52.04 & 41.48 & 37.58 & 39.32 & 52.32 & 48.56 & 49.92 & 47.70 & 25.74 & 33.24\\
    OTE-MTL (ours) & 64.54 & 55.57 & \textbf{59.67}\textsuperscript{\ddag} & 54.18 & 45.20 & \textbf{48.97}\textsuperscript{\ddag} & 58.16 & 54.02 & \textbf{55.83}\textsuperscript{\ddag} & 48.17 & 42.43 & \textbf{45.05}\textsuperscript{\ddag}\\
    \midrule
    OTE-MTL-Inter & 66.24 & 54.38 & 59.61 & 49.32 & 46.12 & 47.33 & 57.71 & 53.06 & 55.17 & 47.66 & 41.85 & 44.43\\
    OTE-MTL-Concat & 48.79 & 48.28 & 48.46 & 46.88 & 42.61 & 44.53 & 52.55 & 48.03 & 50.09 & 46.81 & 38.46 & 42.14\\
    OTE-MTL-Unified & 51.19 & 44.65 & 47.64 & 40.32 & 34.38 & 37.01 & 48.52 & 40.30 & 43.85 & 37.42 & 34.17 & 35.54\\
    OTE-MTL-Collapsed & 45.38 & 36.26 & 40.19 & 32.55 & 29.52 & 30.68 & 37.86 & 33.06 & 35.19 & 32.56 & 27.23 & 29.60\\
    \bottomrule
    \end{tabular}
    }
    \caption{Quantitative evaluation results (\%). Results of models with marker \textsuperscript{*} are reported on datasets without OOTs. Results of models with marker \textsuperscript{\dag} are directly cited from~\citet{peng2019knowing}. F1 measures in \textbf{bold} are the best performing numbers on each dataset. F1 measures with marker \textsuperscript{\ddag} are significantly better than other numbers on each dataset with paired t-test ($p<$  0.01).}
    \label{tab2}
\end{table*}

\subsection{Baselines and Variants}

To perform a systematic comparison, we introduce a variety of baselines, which can be classified into two groups, i.e., pipeline methods proposed in \newcite{peng2019knowing} and joint methods we adapted from previous aspect-opinion co-extraction systems based on our framework \textbf{OTE-MTL}.

First, we list the baselines with a pipeline structure. (1) \textbf{Pipeline}~\cite{peng2019knowing} decomposes triplet extraction to two stages: stage one for predicting unified aspect-sentiment and opinion tags, while stage two for pairing the two results from stage one. We further include three models adjusted in accordance with Pipeline: (2) \textbf{Unified+}~\cite{li2019unified} is a typical aspect-sentiment pair extraction system, in which the unified tagging scheme is used. (3) \textbf{RENANTE+}~\cite{dai2019neural} is originally an aspect-opinion co-extraction system in a weakly-supervised manner. (4) \textbf{CMLA+}~\cite{wang2017coupled} is an aspect-opinion co-extraction system modelling the interaction between the aspects and opinions. Additionally, we adapt two extra baseline models to the multi-task leaning, resulting in: (5) \textbf{CMLA-MTL} and (6) \textbf{HAST-MTL}~\cite{li2018aspect}, which are extended from existing state-of-the-art aspect-opinion co-extraction systems.  

We also propose a list of variants of our proposed OTE-MTL framework to examine the efficacy of different components in it. (a) \textbf{OTE-MTL-Inter} feeds the prediction of aspects and opinions to the biaffine scorer by imposing tag embedding and concatenating tag embeddings to the input of the scorer. (b) \textbf{OTE-MTL-Concat} replaces the biaffine scorer with an activated linear layer applied on the concatenated vectors of aspect and opinion representations. (c) \textbf{OTE-MTL-Unified} uses unified aspect-sentiment tagging scheme and degrades the biaffine scorer to a binary pair classifier, which is similar to Pipeline but is jointly trained. (d) \textbf{OTE-MTL-Collapsed} combines the aspect and opinion tagging components into one single module via a collapsed tag set \{\texttt{B-AP}, \texttt{I-AP}, \texttt{B-OP}, \texttt{B-OP}, \texttt{O}\}, thus is forced to account for the constraint that aspects and opinions would never overlap.
\section{Results and Analysis}

\subsection{Quantitative Evaluation}

\noindent \textbf{Comparison with Baselines.} The results in comparison with baselines are shown in Table~\ref{tab2}, both on datasets with and without OOTs for a fair comparison. Our propose model OTE-MTL consistently outperforms all state-of-the-art baselines on all datasets with and without OOTs. Thus, we conclude OTE-MTL is effective in dealing with opinion triplet extraction task.

We observe that the results of OTE-MTL on datasets without OOTs are generally better than those with OOTs except for \textsc{Laptop14}, implying that datasets without OOTs is comparably simpler and easier to achieve a good performance. Hence, we believe that overlapping cases bring challenges and can be partly addressed via triplet interaction modelling. Nevertheless, CMLA+ presents a worse performance in contrast to superior performance produced by CMLA-MTL. This fact suggests that, through decoupling aspect and sentiment predictions and puting them under the multi-task learning framework, the model can be enhanced and gain better results.     

\noindent \textbf{Comparison with Variants.} The comparison with variants of OTE-MTL shown in Table~\ref{tab2} aims to verify the effectiveness of different components of OTE-MTL. As a whole, OTE-MTL surpasses all its variants. Specifically, OTE-MTL is slightly better than OTE-MTL-Inter, however, OTE-MTL exceeds other variants by large margins. 

Rather than implicitly modelling the interaction between tagging and sentiment dependency parsing, OTE-MTL-Inter explicitly feeds emebddings of predicted tags to the biaffine scorer. It gets an inferior performance. We conjecture the reason lies in the latent error propagation when tags are partially wrong, therefore hinting implicit modelling is a promising choice. The failure of OTE-MTL-Concat, which cannot model priors, supports the idea of leveraging biaffine scorer as word-level sentiment dependency parser. The result of OTE-MTL-Unified indicates that coupling aspect and sentiment extraction is suboptimal. Furthermore, we use OTE-MTL-Collapsed to account for non-overlap constraint of aspects and opinions, however, it obtains unexpectedly poor results. A possible explanation is that simultaneously collapsing aspect and opinion representations into one space may cause limited capacity for expressiveness.

\subsection{Qualitative Evaluation}

\begin{table*}[ht]
    \centering
    \resizebox{\textwidth}{!}{
    \begin{tabular}{cccc}
    \toprule
    Case & Ground truth & OTE-MTL-Unified & OTE-MTL\\
    \midrule
    \makecell[c]{Great food but the \\service was dreadful !} & \makecell[c]{[(food, Great, \texttt{POS}), \\(service, dreadful, \texttt{NEG})]} & \makecell[c]{[(food, Great, \texttt{POS}), \\(service, dreadful, \texttt{NEG})]} & \makecell[c]{[(food, Great, \texttt{POS}), \\(service, dreadful, \texttt{NEG})]}\\
    \midrule
    \makecell[c]{The atmosphere is attractive , \\but a little uncomfortable .} & \makecell[c]{[(atmosphere, attractive, \texttt{POS}), \\(atmosphere, uncomfortable, \texttt{NEG})]} & \makecell[c]{[(atmosphere, attractive, \texttt{POS}), \\(atmosphere, uncomfortable, \texttt{POS}\textsuperscript{\xmark})]} & \makecell[c]{[(atmosphere, attractive, \texttt{POS}), \\(atmosphere, uncomfortable, \texttt{NEG})]}\\
    \midrule
    \makecell[c]{I am pleased with the fast log on , \\speedy WiFi connection and \\the long battery life .} & \makecell[c]{[(log on, fast, \texttt{POS}), \\(WiFi connection, speedy, \texttt{POS}), \\(battery life, long, \texttt{POS}), \\(log on, pleased, \texttt{POS}), \\(WiFi connection, pleased, \texttt{POS}), \\(battery life, pleased, \texttt{POS})]} & \makecell[c]{[(log\textsuperscript{\xmark}, fast, \texttt{POS}), \\(WiFi connection, speedy, \texttt{POS}), \\(battery life, long, \texttt{POS}), \\(log\textsuperscript{\xmark}, pleased, \texttt{POS}), \\()\textsuperscript{\xmark}, \\()\textsuperscript{\xmark}]} & \makecell[c]{[(log\textsuperscript{\xmark}, fast, \texttt{POS}), \\(WiFi connection, speedy, \texttt{POS}), \\(battery life, long, \texttt{POS}), \\(log\textsuperscript{\xmark}, pleased, \texttt{POS}), \\(WiFi\textsuperscript{\xmark}, pleased, \texttt{POS}), \\()\textsuperscript{\xmark}]} \\
    \bottomrule
    \end{tabular}
    }
    \caption{Case study. Marker \textsuperscript{\xmark} indicates incorrect predictions.}
    \label{tab3}
\end{table*}

\noindent \textbf{Case Study.} To understand in what way our framework overwhelms the other unified tagging-based approaches, we perform a case study on three representative examples from test sets, as displayed in Table~\ref{tab3}. 

We notice that both OTE-MTL-Unified and OTE-MTL are working well for the first case which involves no overlapping. Nonetheless, OTE-MTL-Unified performs less well when faced with the second sample which contains aspect overlapped triplets and requires triplet interaction modelling. This case also shows conflicting opinions to an aspect~\cite{tan-etal-2019-recognizing}, which is not covered by the training set but exists in real-world applications. It cannot be coped with by coupled aspect-sentiment tags since a tag should not have diverse sentiments. Thus decoupling sentiments from aspect tags is necessary. In the third example with long-range dependency, both aspect overlap and opinion overlap exist. For this case, OTE-MTL is not strong enough to make all correct predictions, but still seems to work better than OTE-MTL-Unified.

\noindent \textbf{Error Analysis.} To further find out the strengths and limitations of OTE-MTL, we conduct a detailed analysis of false positives (extracted by the system but not existing in ground truth) and false negatives (not extracted by the system but existing in ground truth) on \textsc{Rest14}. For false positives, we categorize them into four classes: false aspect, false opinion, false sentiment, and other (mixed) case. For false negatives, we divide them according to categories of overlap (i.e., aspect overlapped, opinion overlapped, normal). 

\begin{figure}[ht]
    \centering\includegraphics[width=.47\textwidth]{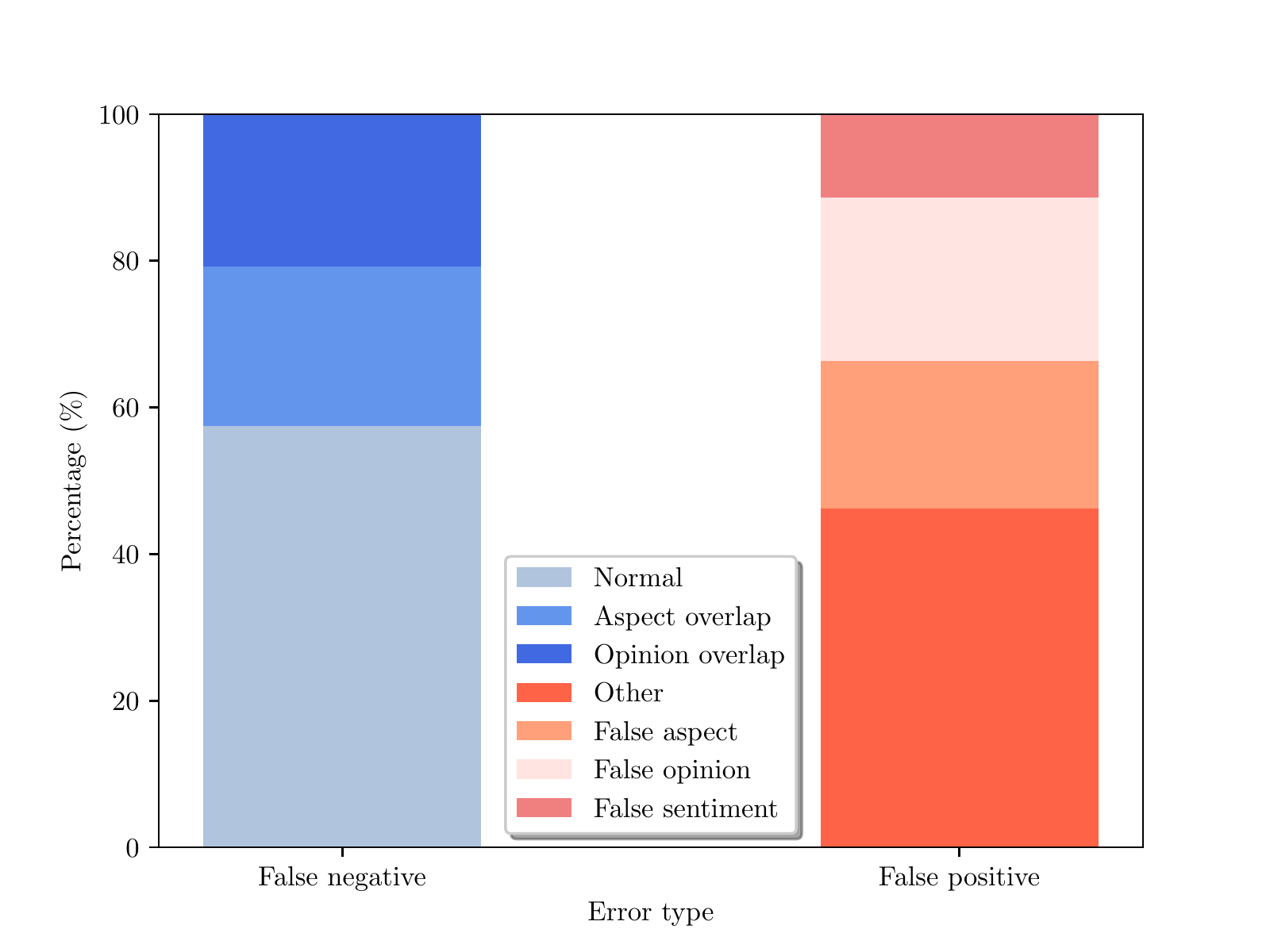}
    \caption{Components of false positives and false negatives.}
    \label{fig4}
\end{figure}

Figure~\ref{fig4} shows the analysis result. False positives are largely triggered by only one false element, especially, aspect or opinion, of an extracted triplet, motivating us to develop more robust span detection algorithms. In addition, the circumstance might also reflect that exact match is not an ideal metric when systems are evaluated, since minor discrepancy in a span may be harmless for opinion interpretation in practice, as we could observe in Table~\ref{tab3}. Likewise, from Figure~\ref{fig4}, we posit that overlapping cases are still non-trivial to solve given they have almost taken half of the false negatives.

\section{Related Work}

\subsection{Aspect-based Sentiment Analysis}

Our work falls in the broad scope of ABSA. As we have previously discussed, there are two types of approaches in ABSA: aspect-sentiment pair extraction that concentrates on collaboratively detecting aspects and attached sentiment orientations~\cite{li2019unified,he2019interactive,luo2019doer,hu2019open}, and aspect-opinion co-extraction that tends to co-extract aspects and opinions~\cite{wang2017coupled,li2018aspect}. Alternatively, ABSA is also formulated as determining sentiment polarity of a given aspect in a sentence~\cite{jiang2011target,dong2014adaptive,tang2016effective,tang2016aspect,li2018transformation,zhang2019aspect}, which is inflexible for practical use since aspects are not naturally accessible.

In this paper, we unify the aspect-sentiment pair extraction and aspect-opinion co-extraction, and formulate them as a triplet extraction problem. Our work is also aimed at addressing several issues in \citet{peng2019knowing}, as discussed in the Introduction Section.

\subsection{Triplet Extraction-based Task}

Other than ABSA, a majority of triplet extraction-based tasks lies in the area of natural language processing. For example, Joint Entity and Relation Extraction (JERE) aims at detecting a pair of entity mentions in a sentence and predicting relation between the two. Approaches to JERE can be sorted into four streams: pipeline-based, table filling-based~\cite{miwa2014modeling,bekoulis2018joint,fu2019graphrel}, tagging-based~\cite{zheng2017joint}, and encoder decoder-based~\cite{zeng2018extracting}. Our work is motivated by table filling methods in \citet{miwa2014modeling} and \citet{bekoulis2018joint}. We decompose triplet extraction to three subtasks, in which word-level sentiment dependency parsing can actually be viewed as a table filling problem, and solve them jointly in a multi-task learning framework.
\section{Conclusions and Future Work}

Our work put forwards an opinion triplet extraction perspective for aspect-based sentiment analysis. Existing works that are applicable to opinion triplet extraction have been shown insufficient, owing to the use of unified aspect-sentiment tagging scheme and ignorance of the interaction between elements in the triplet. Thus, we propose a multi-task learning framework to address the limitations by highlighting the uses of joint training, decoupled aspect and sentiment prediction, and regularization among correlated tasks during learning. Experimental results verify the effectiveness of our framework in comparison with a wide range of strong baselines. Comparison results with different variants of the proposed framework signify the necessity of the core components in the framework.

Based on the observations from a case study and error analysis, we plan to carry out further research in the following aspects: (1) more robust taggers for aspect and opinion extraction, (2) more flexible evaluation metric for triplet extraction, and (3) more mighty triplet interaction mechanism (e.g., encoder decoder structure). 

\section*{Acknowledgments}

This work is supported by The National Key Research and Development Program of China (grant No. 2018YFC0831704) and Natural Science Foundation of China (grant No. U1636203, U1736103).

\bibliography{anthology,emnlp2020}
\bibliographystyle{acl_natbib}




\end{document}